%% file: main.tex
\documentclass[10pt,twocolumn,letterpaper]{article}

\usepackage{iccv}
\usepackage{times}
\usepackage{epsfig}
\usepackage{graphicx}
\usepackage{amsmath}
\usepackage{amssymb}
\usepackage{bm}
\usepackage{microtype}
\usepackage{subfigure}
\usepackage{booktabs}
\usepackage{caption}
\usepackage{multirow}
\usepackage{makecell}
\usepackage{enumitem}
\usepackage[numbers,sort,compress]{natbib}
\usepackage[accsupp]{axessibility}  
\renewcommand{\captionlabelfont}{\small}

\usepackage[pagebackref=true,breaklinks=true,letterpaper=true,colorlinks,bookmarks=false]{hyperref}

\usepackage{xcolor}
\definecolor{deepred}{HTML}{940000}
\hypersetup{linkcolor=deepred}
\hypersetup{citecolor=[rgb]{0.4,0.15,0.95}}
\usepackage{url}

\usepackage{colortbl}
\definecolor{Gray}{gray}{0.94}

\newlength\savewidth\newcommand\shline{\noalign{\global\savewidth\arrayrulewidth
\global\arrayrulewidth 1pt}\hline\noalign{\global\arrayrulewidth\savewidth}}

\iccvfinalcopy 

\ificcvfinal\fi

\begin{document}

\title{\vspace{-6.25mm}Pairwise Similarity Learning is SimPLE}

\author{\fontsize{11.25pt}{\baselineskip}\selectfont Yandong Wen\textsuperscript{1,*},~~~~Weiyang Liu\textsuperscript{1,2,*},~~~~Yao Feng\textsuperscript{1},~~~~Bhiksha Raj\textsuperscript{3,4},~~~~Rita Singh\textsuperscript{3}\\\fontsize{11.25pt}{\baselineskip}\selectfont Adrian Weller\textsuperscript{2,5},~~~~Michael J. Black\textsuperscript{1},~~~~Bernhard Schölkopf\textsuperscript{1}\\[0.75mm]
\fontsize{10.5pt}{\baselineskip}\selectfont\textsuperscript{1}Max Planck Institute for Intelligent Systems,~~~~\textsuperscript{2}University of Cambridge,~~~~\textsuperscript{3}Carnegie Mellon University,
\\\fontsize{10.5pt}{\baselineskip}\selectfont\textsuperscript{4}Mohamed bin Zayed University of Artificial Intelligence,~~~~\textsuperscript{5}The Alan Turing Institute,~~~~\textsuperscript{*}Equal contribution
}

\maketitle
\ificcvfinal\thispagestyle{empty}\fi

\begin{abstract}
   In this paper, we focus on a general yet important learning problem, pairwise similarity learning~(PSL). PSL subsumes a wide range of important applications, such as open-set face recognition, speaker verification, image retrieval and person re-identification. The goal of PSL is to learn a pairwise similarity function assigning a higher similarity score to positive pairs (i.e., a pair of samples with the same label) than to negative pairs (i.e., a pair of samples with different label). We start by identifying a key desideratum for PSL, and then discuss how existing methods can achieve this desideratum. We then propose a surprisingly simple proxy-free method, called SimPLE, which requires neither feature/proxy normalization nor angular margin and yet is able to generalize well in open-set recognition. We apply the proposed method to three challenging PSL tasks: open-set face recognition, image retrieval and speaker verification. Comprehensive experimental results on large-scale benchmarks show that our method performs significantly better than current state-of-the-art methods. Our project page is available at \url{simple.is.tue.mpg.de}.
\end{abstract}

\vspace{-1mm}
\section{Introduction}

How to learn discriminative representations is arguably one of the most fundamental and important problems in computer vision, speech processing and natural language processing. For closed-set classification (\eg, image recognition), it is sufficient to learn class-separable representations as the goal is to infer the label of the input sample. However, for open-set recognition problems such as face recognition~\cite{geng2020recent}, speaker verification~\cite{bimbot2004tutorial}, person re-identification~\cite{ye2021deep} and image retrieval~\cite{datta2008image}, learning class-separable representations is not enough, because the goal becomes learning a similarity function that separates positive and negative pairs well. We study a general problem that is abstracted from these applications -- 
pairwise similarity learning~(PSL).

\begin{figure}[t]
  \centering  
  \includegraphics[width=0.45\textwidth]{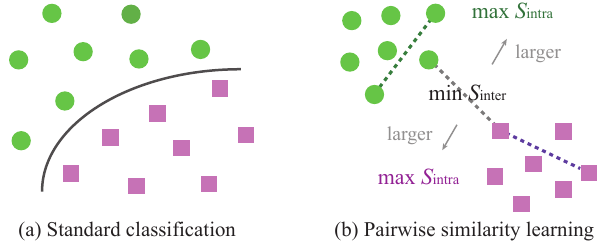}
  \vspace{-1.2mm}
  \caption{\small Comparison between classification and PSL.}
  \vspace{-2.1mm}
  \label{fig:teaser}
\end{figure}

PSL aims to learn a pairwise similarity function such that minimal intra-class similarity is larger than maximal inter-class similarity (or in other words, maximal intra-class distance is smaller than minimal inter-class distance). When this criterion is satisfied, one can easily find a universal threshold that perfectly separates arbitrary positive and negative sample pairs. This property suggests that (i) perfect verification can be achieved and (ii) labels can be fully recovered by simple hierarchical clustering. Compared to classification, PSL presents a more challenging problem of learning large-margin representations, as illustrated by Figure~\ref{fig:teaser}. 

PSL can be viewed as a generalization of deep metric learning~(DML). While DML requires the dissimilarity function to be a distance metric that satisfies non-negativity and the triangle inequality, PSL does not necessarily need to follow these criteria. For example, \cite{liu2017sphereface,wang2018additive,wang2017normface,deng2019arcface} learn a cosine similarity that separates positive and negative pairs. 

\subsection{Desideratum for Pairwise Similarity Learning}

We start by formally describing the desideratum of PSL. PSL seeks to learn a pairwise similarity function $\mathcal{S}(\bm{x}_1,\bm{x}_2)$ that is typically symmetric (\ie, $\mathcal{S}(\bm{x}_1,\bm{x}_2)=\mathcal{S}(\bm{x}_2,\bm{x}_1)$). The desired pairwise similarity function needs to always satisfy the following inequality: $\mathcal{S}(\bm{x}_i,\bm{x}_j) > \mathcal{S}(\bm{x}_p,\bm{x}_q)$ where $\bm{x}_i,\bm{x}_j$ denote an arbitrary pair of samples with the same label, and $\bm{x}_p,\bm{x}_q$ denote an arbitrary pair with different labels. This inequality implies that no negative pair has a larger similarity score than positive pairs. With a labeled dataset, we can further interpret the criterion as
\begin{equation}\label{eq:des2}
\small
\underbrace{\min_{k,i\neq j}\mathcal{S}(\bm{x}_{i}^{[k]},\bm{x}_j^{[k]})}_{\text{Minimal intra-class similarity score}} > \underbrace{\max_{m\neq n,p,q} \mathcal{S}(\bm{x}_p^{[m]},\bm{x}_q^{[n]})}_{\text{Maximal inter-class similarity score}}
\end{equation}
where $\bm{x}_{i}^{[k]}$ denotes the $i$-th sample in the $k$-th class. Normally, there are two ways to parameterize the similarity function. A naive way is to directly parameterize the similarity function as a neural network $\bm{\theta}$, resulting in $\mathcal{S}_{\bm{\theta}}(\bm{x}_i,\bm{x}_j)$. 
However, this parameterization is not scalable for inference, since obtaining all the pairwise similarity scores for $N$ samples
requires  network inference with complexity $\mathcal{O}(N^2)$. 
A more sensible way is to, instead, parameterize the similarity score as $\mathcal{S}(f_{\bm{\theta}}(\bm{x}_i),f_{\bm{\theta}}(\bm{x}_j))$, where $\mathcal{S}$ is usually a simple and efficient similarity measure (\eg, cosine similarity) and $f_{\bm{\theta}}$ is a neural feature encoder parameterized by $\bm{\theta}$. Such a parameterization only requires $\mathcal{O}(N)$ time for network inference and $\mathcal{O}(N^2)$ time for simple pairwise similarity function evaluation. Current PSL methods really boil down to learning a feature encoder that can achieve Eq.~\ref{eq:des2}.

The criterion of Eq.~\ref{eq:des2} essentially suggests that simple clustering of features leads to perfect classification. Unlike classification problems seeking separable feature representations, PSL aims at large-margin features such that the labels can be recovered by hierarchical clustering. Bearing the desideratum in mind, we first examine how existing PSL methods approach this goal, and then propose a PSL framework that is surprisingly simple yet effective to achieve this.

\subsection{Taxonomy of Pairwise Similarity Learning}

Towards such a desideratum, there are currently 
two main types of method:
proxy-based PSL (\eg, \cite{liu2017sphereface,deng2019arcface,wang2018additive,wang2018cosface}) and proxy-free PSL (\eg, \cite{hadsell2006dimensionality,schroff2015facenet,taigman2014deepface}). Proxy-based PSL utilizes an intermediate parametric sample to serve as a proxy for a group of samples (typically one proxy for one class), which has been shown to benefit convergence and training stability. However, these advantages also come at a price in the sense that it is more difficult for proxy-based PSL to achieve the desideratum. How to achieve Eq.~\ref{eq:des2} with the presence of proxies is highly nontrivial and usually requires additional design to the loss function~\cite{liu2016large,liu2017sphereface}. Typical examples of proxy-free PSL include contrastive loss~\cite{chopra2005learning,hadsell2006dimensionality} and triplet loss~\cite{weinberger2009distance}, where no proxies are used during training. Although proxy-free PSL can easily use Eq.~\ref{eq:des2} as the training target, how to construct pairs or triplets becomes especially crucial for convergence and generalization. Hard sample mining matters significantly for performance \cite{wu2017sampling}. Because Eq.~\ref{eq:des2} is generally intractable to achieve for large training sets, the key difference between proxy-based PSL and proxy-free PSL originates from how they approximate this criterion. Proxy-based PSL achieves Eq.~\ref{eq:des2} by crafting a relationship between samples and proxies. Proxy-free PSL implements Eq.~\ref{eq:des2} by sampling a few representative intra-class and inter-class sample pairs, rather than enumerating all the possible pairs. Therefore, how these representative pairs are selected plays a crucial role in determining whether Eq.~\ref{eq:des2} can be effectively achieved.

Categorization of PSL can also be made from the perspective of how the similarity scores between different pairs interact with each other during optimization~\cite{wen2021sphereface2}. Specifically, if the training involves comparing the similarity scores between different pairs, then we call it triplet-based learning. Typical examples include triplet loss~\cite{weinberger2009distance,schroff2015facenet} and almost all the margin-based softmax cross-entropy losses~\cite{liu2017sphereface,deng2019arcface,wang2018additive,wang2018cosface,meng2021magface,kim2022adaface}. In contrast, if the training directly compares the pair similarity scores to a universal value, then we call it pair-based learning. Examples include contrastive loss~\cite{chopra2005learning} and binary cross-entropy~\cite{wen2021sphereface2}. For downstream tasks that focus on comparing pairs of samples, pair-based learning can be preferable since its training objective is more aligned with the testing scenario.

\begin{table}[t]
	\centering
	\scriptsize
	\newcommand{\tabincell}[2]{\begin{tabular}{@{}#1@{}}#2\end{tabular}}
	\setlength{\tabcolsep}{.9pt}
  \renewcommand{\arraystretch}{1.15}
\renewcommand{\captionlabelfont}{\small}
\begin{tabular}{c|cc|cc} 
  & \multicolumn{2}{c|}{Proxy-based} & \multicolumn{2}{c}{Proxy-free} \\
  & Angular & Non-angular & Angular & Non-angular\\ \shline
Triplet
& 
\begin{tabular}{c}
VGGFace~\cite{parkhi2015deep} \\
Triplet~\cite{ming2017simple} \\
SphereFace~\cite{liu2017sphereface} \\
NormFace~\cite{wang2017normface} \\
CosFace~\cite{wang2018cosface,wang2018additive} \\
ArcFace~\cite{deng2019arcface} \\
SoftTriple~\cite{qian2019softtriple}\\
Circle~Loss~\cite{sun2020circle} \\
HUG~\cite{Liu2023HUG}
\end{tabular}
&
\begin{tabular}{c}
DeepID~\cite{sun2014deep}\\
DeepFace~\cite{taigman2014deepface}\\
DeepID2\textsuperscript{*}~\cite{sun2014deep2}\\
L-Softmax~\cite{liu2016large}\\
Center Loss\textsuperscript{*}~\cite{wen2016discriminative}\\
Proxy NCA~\cite{movshovitz2017no}\\
Proxy-Anchor~\cite{kim2020proxy}
\end{tabular}
& 
\begin{tabular}{c}
FaceNet~\cite{schroff2015facenet} \\
Angular Loss~\cite{wang2017deep} \\
Tuplet~\cite{yu2019deep}\\
SupCon~\cite{khosla2020supervised}\\
Smooth-AP~\cite{brown2020smooth}\\
HUG~\cite{Liu2023HUG}
\end{tabular}
&  
\begin{tabular}{c}
Triplet Loss~\cite{weinberger2009distance} \\
N-pair~\cite{sohn2016improved} \\
LiftedStruct~\cite{song2016deep} \\
InfoNCE~\cite{oord2018representation}\\
Log-ratio~\cite{kim2019deep}\\
Ranked List~\cite{wang2019ranked}\\
SNR~\cite{yuan2019signal}
\end{tabular}
\\\hline
\multirow{1}{*}[0pt]{Pair} & 
\begin{tabular}{c}
SphereFace2~\cite{wen2021sphereface2}
\end{tabular}
&
\begin{tabular}{c}
BCE~\cite{khan2020unsupervised,kuroyanagi2021anomalous}\\
Center Loss\textsuperscript{*}~\cite{wen2016discriminative}
\end{tabular}
&
\begin{tabular}{c}
AMC-Loss~\cite{choi2020amc}\\
RBM~\cite{nair2010rectified}\\
\end{tabular}
&
\begin{tabular}{c}
Siamese~\cite{chopra2005learning,hadsell2006dimensionality}\\
DeepID2\textsuperscript{*}~\cite{sun2014deep2}\\
Multi-sim~\cite{wang2019multi}\\
SNR~\cite{yuan2019signal}\\
\textbf{SimPLE}
\end{tabular}
\\
\specialrule{0em}{0pt}{-4pt}
\end{tabular}
\caption{\small Taxonomy of some representative PSL methods. * indicates that the method has hybrid components.}
\label{tab:taxonomy}
\vspace{-2mm}
\end{table}

There are also several similarity functions that are widely adopted in PSL: angular similarity~\cite{schroff2015facenet,liu2017sphereface,deng2019arcface,wang2018additive,wang2018cosface,wang2017deep}, inner product~\cite{sun2014deep,sun2014deep2} and Euclidean
distance~\cite{chopra2005learning,hadsell2006dimensionality}. Angular similarity has become a \emph{de facto} choice in open-set recognition, since it can effectively avoid degenerate solutions in triplet-based learning~\cite{schroff2015facenet,Liu2021SphereFaceR} and also help to incorporate angular margin for softmax cross-entropy losses~\cite{liu2017sphereface,wang2018cosface,wang2018additive,deng2019arcface,wen2021sphereface2,Liu2021SphereFaceR}. We summarize a taxonomy for some representative PSL methods in Table~\ref{tab:taxonomy}.

\subsection{Motivation and Contribution}

Looking into Eq.~\ref{eq:des2} for PSL, we can observe a few characteristics: (1) similarity is only computed between samples and no proxies are involved; (2) there exists a universal threshold that separates intra-class similarity score and inter-class similarity score. The two observations suggest that pair-based proxy-free learning is best aligned with the desideratum. Despite the perfect alignment between the training target of pair-based proxy-free learning and the desideratum, this category remains largely unexplored and existing methods from it are not particularly competitive. Some natural questions arise: \emph{Why don't pair-based proxy-free PSL methods work as well as expected?
Can we realize the full potential for this type of method?} Driven by these questions, our paper studies pair-based proxy-free learning and develops a working algorithm for this approach.

To this end, we first challenge the necessity of a few \emph{de facto} components in state-of-the-art PSL methods, such as angular similarity~\cite{schroff2015facenet,wang2017normface,liu2017deep,liu2017sphereface} and angular margin~\cite{liu2017sphereface}, and then propose a surprisingly simple yet effective pair-based proxy-free PSL framework, dubbed \emph{SimPLE}, where neither angular similarity nor margin is needed. Our major contributions can be summarized as follows:

\vspace{1mm}

\begin{itemize}[leftmargin=*,nosep]
\setlength\itemsep{0.4em}
    \item We rethink the desideratum of pairwise similarity learning, which effectively subsumes many important applications. We identify that pair-based proxy-free learning is most aligned with such a desideratum.
    \item We challenge a few dominant components in current PSL methods (\eg, angular similarity and margin), and find them unnecessary in the pair-based proxy-free regime.
    \item We propose SimPLE, a surprisingly simple yet effective pair-based proxy-free learning framework that is designed directly based on the desideratum of PSL.
    \item Most importantly, we show that SimPLE can easily achieve state-of-the-art performance on open-set face recognition, image retrieval, and speaker verification. We note that this is the first time that a PSL method achieves 
    state-of-the-art performance without the help of angular similarity and margin in open-set face recognition.
\end{itemize}

\section{Rethinking Pairwise Similarity Learning}\label{sect:rethink}

We start by examining how different types of PSL methods achieve Eq.~\ref{eq:des2}. Since proxy-based PSL models the relationship between samples and proxies, it approximates Eq.~\ref{eq:des2} through the constraint embedded in the similarity function. Specifically, we consider a two-class scenario. We have samples $\bm{x}_i$ and $\bm{x}_j$ from the first class constitute the minimal intra-class similarity. $\bm{x}_k$ (class 1) and $\bm{z}_k$ (class 2) yield the maximal inter-class similarity. Then PSL's desideratum requires us to have $\mathcal{S}(\tilde{\bm{x}}_i,\tilde{\bm{x}}_j)>\mathcal{S}(\tilde{\bm{x}}_k,\tilde{\bm{z}}_k)$ where we define $\tilde{\bm{x}}=f_{\bm{\theta}}(\bm{x})$ for notation convenience. For proxy-based PSL to achieve this inequality, we first consider a triangular inequality for the similarity score function:
\begin{equation}
\small
\begin{aligned}
        \mathcal{S}(\bm{v}_1,\bm{v}_3) - \mathcal{S}(\bm{v}_2,\bm{v}_3) \geq \mathcal{S}(\bm{v}_1,\bm{v}_2) \geq \mathcal{S}(\bm{v}_1,\bm{v}_3) + \mathcal{S}(\bm{v}_2,\bm{v}_3)\nonumber
\end{aligned}
\end{equation}
which is also satisfied by the prominent angular similarity, \ie, $\mathcal{S}(\bm{v}_1,\bm{v}_2)=1-\frac{1}{\pi}\arccos(\frac{\bm{v}_1^\top\bm{v}_2}{\|\bm{v}_1\|\cdot\|\bm{v}_2\|})$. Then we have
\begin{equation}
\small
\begin{aligned}
    &\mathcal{S}(\tilde{\bm{x}}_i,\bm{w}_1) - \mathcal{S}(\tilde{\bm{x}}_j,\bm{w}_1) \geq\mathcal{S}(\tilde{\bm{x}}_i,\tilde{\bm{x}}_j) \\
    &\mathcal{S}(\tilde{\bm{x}}_k,\tilde{\bm{z}}_k)\geq \mathcal{S}(\tilde{\bm{x}}_k,\bm{w}_2) + \mathcal{S}(\tilde{\bm{z}}_k,\bm{w}_2)
\end{aligned}
\end{equation}
which leads to the following sufficient condition for $\mathcal{S}(\tilde{\bm{x}}_i,\tilde{\bm{x}}_j)>\mathcal{S}(\tilde{\bm{x}}_k,\tilde{\bm{z}}_k)$ to hold:
\begin{equation}
\small
\begin{aligned}
    \underbrace{\mathcal{S}(\tilde{\bm{x}}_i,\bm{w}_1)}_{\text{Intra-class similarity}} - \underbrace{\big(\mathcal{S}(\tilde{\bm{x}}_j,\bm{w}_1) + \mathcal{S}(\tilde{\bm{x}}_k,\bm{w}_2) \big)}_{\text{Margin between similarity scores}}>\underbrace{\mathcal{S}(\tilde{\bm{x}}_k,\bm{w}_2)}_{\text{Inter-class similarity}}\nonumber
\end{aligned}
\end{equation}
where $\bm{w}_1$ and $\bm{w}_2$ denote the proxy for class 1 and 2, respectively. For proxy-based PSL to achieve the desideratum, we have to introduce a margin between intra-class and inter-class similarity score. Without the margin, $\mathcal{S}(\tilde{\bm{x}}_i,\bm{w}_1)>\mathcal{S}(\tilde{\bm{x}}_k,\bm{w}_2)$ is the criterion for standard classification and only implies separable features. The triangular inequality indicates that with a proper distance metric being the dissimilarity function, it will be easier for proxy-based PSL to achieve the desideratum in Eq.~\ref{eq:des2}. Therefore, margin is actually indispensable for proxy-based PSL.

Then we discuss why angular similarity is widely adopted in PSL. We consider the softmax cross-entropy loss:
\begin{equation}
\small
\mathcal{L}_{\text{CE}}=\log\big(1+\sum_{i\neq y}\exp(\bm{w}_i^\top\tilde{\bm{x}} - \bm{w}_y^\top\tilde{\bm{x}})\big)
\end{equation}
where $\bm{w}_i$ denotes the $i$-th class proxy (\ie, last-layer classifier) and $\tilde{\bm{x}}$ is the feature ($y$ is the label). We have that

\vspace{-4mm}
\begin{equation}
\small
\lim_{\|\bm{x}\|\rightarrow\infty}\mathcal{L}_{\text{CE}}=\left\{
{\begin{array}{*{20}{l}}
0 & \textnormal{if} & \forall i\neq y,~\bm{w}_y^\top\tilde{\bm{x}}>\bm{w}_i^\top\tilde{\bm{x}}\\
+\infty & \textnormal{if} & \exists i\neq y,~\bm{w}_y^\top\tilde{\bm{x}}<\bm{w}_i^\top\tilde{\bm{x}}
\end{array}} \right.
\end{equation}
which implies that as long as the feature can be classified to the correct class (\ie, features are separable), then the softmax cross-entropy loss can be trivially minimized by increasing the feature norm. In order to eliminate these degenerate solutions, common practice~\cite{liu2017sphereface,wang2017normface,wang2018additive,wang2018cosface,deng2019arcface,liu2021orthogonal} resorts to normalizing both proxy weights and features to a fixed length, leading to the popular angular similarity. One may notice an obvious caveat here -- both angular margin and angular similarity are especially designed for proxy-based learning. Neither is necessary for proxy-free learning.

\begin{figure}[t]
  \centering  
  \includegraphics[width=0.435\textwidth]{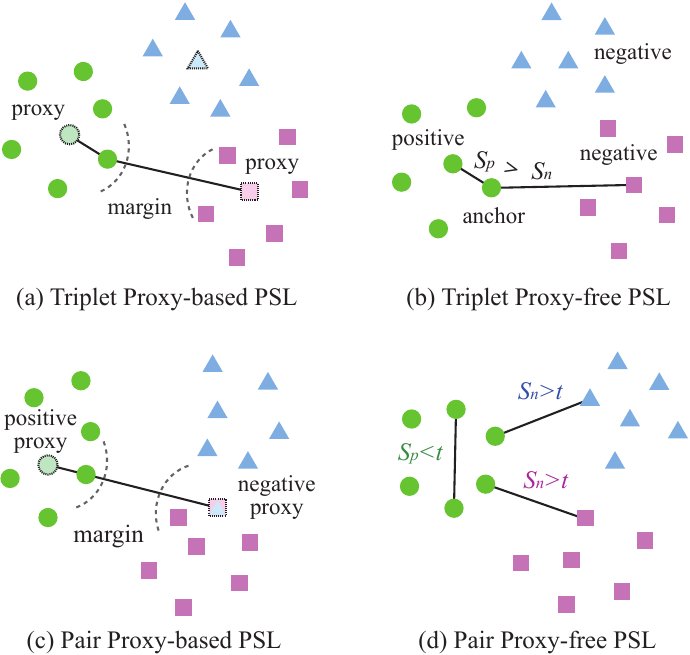}
  \vspace{-0.75mm}
  \caption{\small Comparison of different types of PSL.}
  \vspace{-2.1mm}
  \label{fig:rethink}
\end{figure}

Now we discuss how proxy-free learning approximates the desideratum in Eq.~\ref{eq:des2}. Contrastive loss~\cite{chopra2005learning,hadsell2006dimensionality} and triplet loss~\cite{weinberger2009distance,schroff2015facenet} are arguably the most representative proxy-free PSL methods. Since it is computationally intractable to enumerate all the possible sample pairs or triplets, both contrastive and triplet losses heavily rely on hard sample mining which essentially seeks representative samples to approximate the minimal intra-class and maximal inter-class similarity. Moreover, triplet loss also has similar degenerate solutions as the softmax cross-entropy loss, so it is usually used together with feature normalization~\cite{schroff2015facenet}. One significant difference between triplet-based learning and pair-based learning is the use of a universal threshold. For example, a triplet loss enforces the similarity between an anchor and a positive sample to be larger than the similarity between the anchor and a negative sample. In contrast, pair-based learning (\eg, contrastive loss and SphereFace2~\cite{wen2021sphereface2}) compares both positive and negative pairs to a universal threshold, which inherently draws a consistent decision boundary between positive pairs and negative pairs and is more aligned with PSL's desideratum. 

We give an intuitive comparison of different types of PSL in Figure~\ref{fig:rethink}. The target of pair-based proxy-free PSL is perfectly aligned with the desideratum that minimal intra-class similarity score is larger than maximal inter-class similarity score. Surprisingly, we find that neither angular similarity (\ie, feature/proxy normalization) nor angular margin is necessary. Further, we identify two important aspects that are essential for pair-based proxy-free PSL: (1) pair sampling, which affects how accurately it can approximate the desideratum in Eq.~\ref{eq:des2}; (2) similarity score, which should be consistent across training and testing. In Section~\ref{sect:SimPLE}, we discuss how we use a simple design to address these issues.

\section{An Embarrassingly Simple PSL Framework}\label{sect:SimPLE}

We aim for SimPLE to be as simple as possible without introducing additional assumptions or priors. We formulate pair-based proxy-free PSL as a pair classification problem, which yields the following naive loss formulation:
\begin{equation}\label{eq:SimPLE_naive}
\small
\begin{aligned}
    \mathcal{L}_{\text{n}} =& \mathbb{E}_{\{\tilde{\bm{x}}_1,\tilde{\bm{x}}_2\}\sim \mathcal{D}} \bigg{\{}y_p\cdot\log\big(1+\exp(-\mathcal{S}(\tilde{\bm{x}}_1,\tilde{\bm{x}}_2)-b)\big)+\\[-0.5mm]
    &~~~~~~~~~~~~~~~ (1-y_p) \cdot\log\big(1+\exp(\mathcal{S}(\tilde{\bm{x}}_1,\tilde{\bm{x}}_2)+b)\big)\bigg{\}}
\end{aligned}
\end{equation}
where $y_p=1$ if $\tilde{\bm{x}}_1$ and $\tilde{\bm{x}}_2$ are from the same class, and $y_p=0$ otherwise. This is essentially a binary logistic regression without classifiers (\ie, binary cross entropy). The advantage of such a formulation can be better understood from its decision boundary $\mathcal{S}(\tilde{\bm{x}}_1,\tilde{\bm{x}}_2)+b=0$. When $\mathcal{S}(\tilde{\bm{x}}_1,\tilde{\bm{x}}_2)$ is larger than $-b$, then $\tilde{\bm{x}}_1$ and $\tilde{\bm{x}}_2$ are predicted to the same class. Otherwise, they are predicted as a negative pair.

\begin{figure}[t]
  \centering  
  \includegraphics[width=0.44\textwidth]{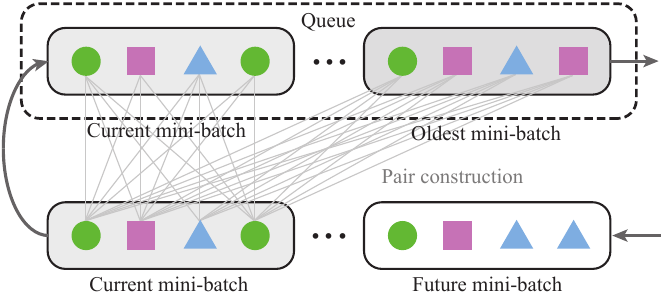}
  \vspace{-0.75mm}
  \caption{\small Illustration of SimPLE's pair construction.}
  \vspace{-2.1mm}
  \label{fig:queue}
\end{figure}

\vspace{2.5mm}

\noindent\textbf{Similarity score}. Cosine similarity (or angular similarity) has been the \emph{de facto} standard in open-set face recognition~\cite{liu2017sphereface,wang2017normface,wang2018additive,wang2018cosface,deng2019arcface}, speaker verification~\cite{wan2018generalized,liu2019large,chung2020defence} and image retrieval~\cite{musgrave2020metric}. Despite its popularity, angular similarity introduces an assumption that features are supposed to be discriminative on the unit hypersphere. However, Eq.~\ref{eq:des2} does not necessitate angular similarity. As long as the similarity score is consistent across training and testing, then we can expect it to generalize well. We start with the simplest case without any assumption -- the inner product as the similarity score: $\mathcal{S}(\tilde{\bm{x}}_1,\tilde{\bm{x}}_2)=\langle \tilde{\bm{x}}_1,\tilde{\bm{x}}_2 \rangle=\|\tilde{\bm{x}}_1\|\cdot\|\tilde{\bm{x}}_2\|\cdot\cos(\theta_{\tilde{\bm{x}}_1,\tilde{\bm{x}}_2}) $. However, the sign of the inner product completely depends on the angle between two features. When the angle is smaller than $\frac{\pi}{2}$, then increasing the similarity can trivially become increasing the feature magnitude. When the angle is larger than $\frac{\pi}{2}$, then decreasing the similarity can also trivially become decreasing the feature magnitude. We find it to be a strong assumption to use $\frac{\pi}{2}$ as the sign boundary. Therefore, we remove such an assumption by adding an angular bias:
\begin{equation}\label{eq:gen_inner}
\small
    \mathcal{S}(\tilde{\bm{x}}_1,\tilde{\bm{x}}_2) = \|\tilde{\bm{x}}_1\|\cdot\|\tilde{\bm{x}}_2\|\cdot\big(\cos(\theta_{\tilde{\bm{x}}_1,\tilde{\bm{x}}_2})-b_{\theta}\big)
\end{equation}
where $b_{\theta}$ is learned directly from data and stays constant during inference. How does this angular bias term differ from the bias term in Eq.~\ref{eq:SimPLE_naive}? We write down the decision boundary for the new similarity functions:
\begin{equation}
\small
\begin{aligned}
\underbrace{\|\tilde{\bm{x}}_1\|\cdot\|\tilde{\bm{x}}_2\|\cdot\cos(\theta_{\tilde{\bm{x}}_1,\tilde{\bm{x}}_2})}_{\text{Inner product similarity}}-\underbrace{\|\tilde{\bm{x}}_1\|\cdot\|\tilde{\bm{x}}_2\|\cdot b_{\theta}}_{\text{Data-dependent bias}}+\underbrace{b}_{\text{Constant bias}}=0\nonumber
\end{aligned}
\end{equation}
which is not equivalent to the decision boundary induced by the inner product similarity. The data-dependent bias serves a different role to the constant bias, and also removes a prescribed assumption in inner product. Our experiments show that removing this assumption is important and leads to consistently better performance. One delicate difference to the angular similarity is that the angular bias is redundant since $\|\tilde{\bm{x}}_1\|\cdot\|\tilde{\bm{x}}_2\|\cdot b_{\theta}$ also becomes some fixed constant and can be trivially merged to $b$ with $\|\tilde{\bm{x}}_1\|=\|\tilde{\bm{x}}_2\|=1$. 

\vspace{2.5mm}

\noindent\textbf{Pair sampling}. How to construct pairs is arguably one of the most important factors in determining the performance of proxy-free learning~\cite{wu2017sampling}. We consider two aspects of pair sampling: pair coverage and pair importance.

Because it is impossible to enumerate all the pair combinations for a large dataset, we seek to enlarge the coverage of pairs. The size of mini-batches also limits the pair coverage. To address this, we maintain a queue of samples encoded by a moving-averaged encoder~\cite{he2020momentum} and then form pairs from samples in the queue. Specifically, we use a first-in-first-out queue where the oldest mini-batch is dequeued as the current mini-batch is enqueued. We denote the size of the mini-batch as $m$ and the size of the queue as $q$. We can form $m\cdot q$ pairs in total. We note that the samples in the queue are encoded by a moving-averaged encoder instead of the original encoder. The moving-averaged encoder is updated by $\bm{\theta}_q\leftarrow \eta \bm{\theta}_q+(1-\eta)\bm{\theta}$ where $\eta$ is the moving average parameter, $\bm{\theta}_q$ are the parameters of the moving-averaged encoder and $\bm{\theta}$ are the parameters of the current encoder that is trained with back-propagation.

\begin{figure}[t]
  \centering  
  \includegraphics[width=0.47\textwidth]{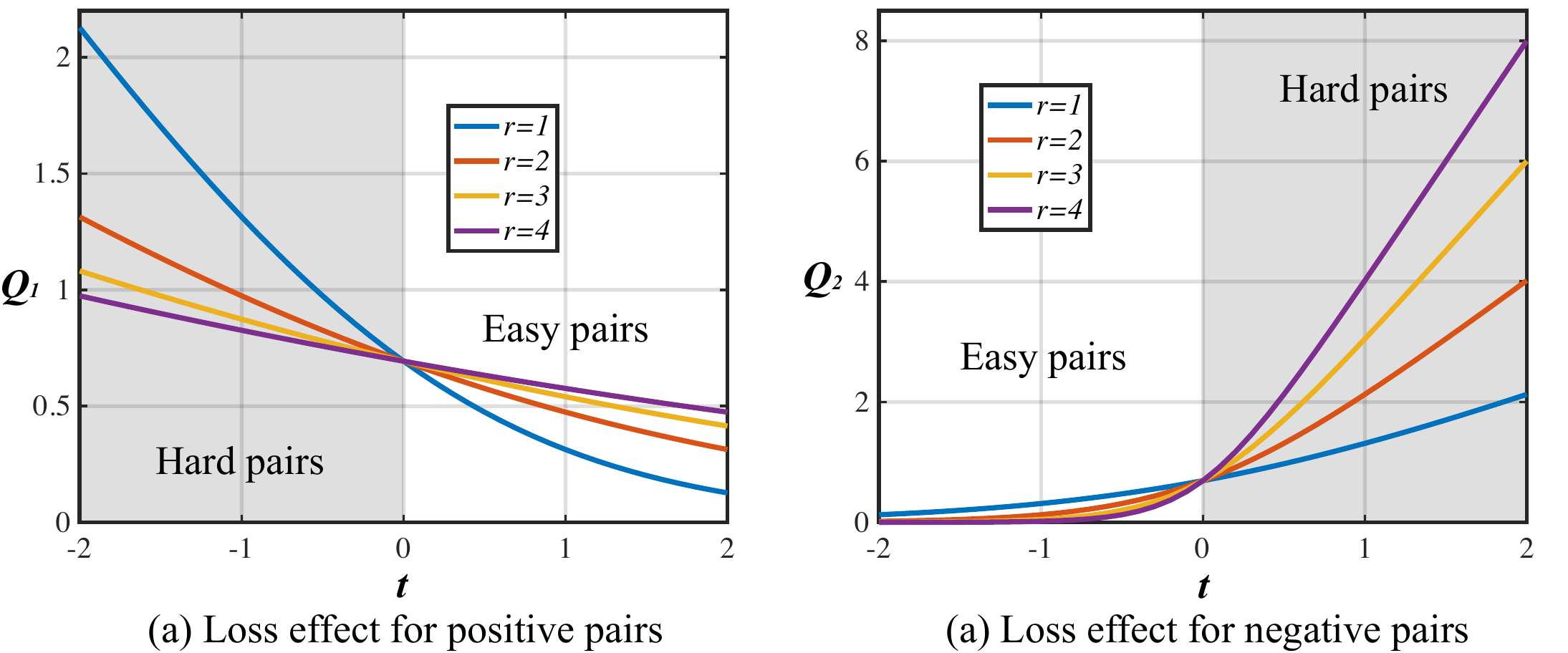}
  \vspace{-1.8mm}
  \caption{\small The effect of $r$ for hard pair mining.}
  \vspace{-3mm}
  \label{fig:hardsample}
\end{figure}

With a sufficient number of pairs, we now consider how to weight them based on their importance. We implement the pair reweighting in the loss function. The way we construct pairs will inevitably result in a highly imbalanced number of positive and negative pairs. To address this problem, we first introduce a weighting hyperparameter to balance the importance of positive and negative pairs, yielding
\begin{equation}\label{eq:lossb}
\small
\begin{aligned}
  &\!\!\mathcal{L}_{\text{b}} = \mathbb{E}_{\{\tilde{\bm{x}}_1,\tilde{\bm{x}}_2\}\sim \mathcal{D}} \bigg{\{}\alpha\cdot y_p\cdot\log\big(1+\exp(-\mathcal{S}(\tilde{\bm{x}}_1,\tilde{\bm{x}}_2)+b)\big)\!\\[-0.5mm]
    &~~~~~+(1-\alpha)\cdot(1-y_p) \cdot\log\big(1+\exp(\mathcal{S}(\tilde{\bm{x}}_1,\tilde{\bm{x}}_2)+b)\big)\bigg{\}}
\end{aligned}
\end{equation}
where $\alpha$ is a hyperparameter for balancing positive and negative pairs. Then we consider the final problem of hard pair mining. We note that hard pair mining is highly nontrivial without angular similarity (\ie, feature and proxy normalization). For example, if we multiply the similarity score by a scaling parameter (\ie, simply replace $\mathcal{S}(\tilde{\bm{x}}_1,\tilde{\bm{x}}_2)$ with $r\cdot \mathcal{S}(\tilde{\bm{x}}_1,\tilde{\bm{x}}_2)$ in Eq.~\ref{eq:lossb}), this parameter will not have the same effect of hard pair mining as SphereFace2~\cite{wen2021sphereface2}. This is because the network can trivially learn to decrease the feature magnitude and $r$ will be compensated by the decreased magnitude, whereas features are normalized in \cite{wen2021sphereface2}. This phenomenon suggests that the effect of hard pair mining within positive and negative pairs tends to cancel out each other in our formulation (without angular similarity).

To address this critical problem, we propose a simple yet novel remedy -- perform hard pair mining in a reverse direction for positive and negative pairs. Specifically, we seek a hyperparameter that simultaneously controls the hard pair mining for both positive and negative pairs. As it gets larger, the loss function focuses more on easy pairs within positive pairs, and at the same time, focuses more on hard pairs within negative pairs. The core idea is that as long as the mining directions are reversed for positive and negative pairs, then their effect will no longer cancel out each other. To this end, we multiply $\frac{1}{r}$ to the similarity score in the loss of positive pairs (instead of $r$), and simultaneously multiply $r$ by the similarity score in the loss of negative pairs. 
We arrive at the final form of the loss function below:
\begin{equation}\label{eq:final_loss}
\small
\begin{aligned}
    &\!\!\mathcal{L}_{\text{f}}\! =\! \mathbb{E}_{\{\tilde{\bm{x}}_1,\tilde{\bm{x}}_2\}\sim \mathcal{D}} \bigg{\{}\!\alpha\!\cdot\! y_p\!\cdot\!\log\!\bigg(\!1\!+\!\exp\big(\!-\!\frac{1}{r}(\mathcal{S}(\tilde{\bm{x}}_1,\tilde{\bm{x}}_2)\!+\!b)\big)\!\bigg)\!\!\\[-0.5mm]
    &~+(1\!-\!\alpha)\cdot(1\!-\!y_p) \cdot\log\bigg(1+\exp\big(r(\mathcal{S}(\tilde{\bm{x}}_1,\tilde{\bm{x}}_2)+b)\big)\!\bigg)\!\bigg{\}}
\end{aligned}
\end{equation}
where $r$ is a hyperparameter that scales the loss curve with respect to the similarity score. Specifically, larger $r$ corresponds to more importance on easy positive pairs and hard negative pairs. We define $Q_1(t)=\log(1+\exp(-t/r))$ and $Q_2(t)=\log(1+\exp(r\cdot t))$, and then plot their curves to illustrate how they achieve hard pair mining of reverse directions. For the function $Q_1(t)$, the loss focuses more on easy pairs as $r$ gets larger. For the function $Q_2$, the loss focuses more on hard samples as $r$ gets larger.

\vspace{2.5mm}

\noindent\textbf{Simplicity and significance of SimPLE}. With similarity score, pair coverage and pair importance taken into account, we end up with a surprisingly simple formulation in Eq.~\ref{eq:final_loss} which only requires simple modifications from standard binary cross-entropy. Most importantly, SimPLE completely drops the dependency on angular similarity and margin while still achieving state-of-the-art performance on almost all open-set recognition problems. We believe this method is 
significant since it opens up new possibilities for PSL and also demonstrates that angular similarity and margin are no longer requisite to achieve state-of-the-art performance.

\section{Discussions and Insights}

\noindent\textbf{SimPLE closes the training-testing gap}. One of the most challenging problems in open-set recognition is the gap between training and testing. Almost all previous innovations were made towards bridging this gap. For example, angular margin~\cite{liu2017sphereface,wang2018cosface,wang2018additive,deng2019arcface,wen2021sphereface2,Liu2021SphereFaceR} is widely adopted in proxy-based learning such that the training target can be closer to the testing scenario. Recently, SphereFace2 \cite{wen2021sphereface2} was proposed to further bridge this gap by switching from triplet-based learning to pair-based learning, because only pair comparison is performed during testing. However, the use of proxies still prevents SphereFace2 from closing this gap, and moreover, SphereFace2 remains heavily dependent on angular similarity and margin. Our work can actually be viewed as a novel proxy-free generalization of SphereFace2. By dropping the use of class proxies, angular similarity and margin, SimPLE takes one step further towards closing the gap between training and testing in open-set recognition.

\vspace{2.5mm}

\noindent\textbf{SimPLE as a general framework}. SimPLE gives a simple yet working variant for pair-based proxy-free learning, but more importantly, SimPLE identifies a few critical design aspects (\eg, similarity score, pair coverage, pair importance) to achieve PSL's desideratum and opens new possibilities. For example, the optimal similarity score is yet to be designed and how to effectively incorporate hard pair mining without the use of angular similarity remains an open problem. Solving any of these open problems might easily lead to better loss functions in pair-based proxy-free learning.



\input{experiments.tex}

\section{Related Work and Concluding Remarks}

How to learn discriminative representations has been a shared goal of multiple lines of research. We conclude our paper by discussing some highly related work.

\vspace{2.1mm}

\noindent\textbf{Contrastive learning}. There has been a rapidly growing interest~\cite{oord2018representation,tian2020contrastive,grill2020bootstrap,he2020momentum,chen2020simple,chen2021exploring} in learning representations with instance contrast. The core idea is to view a sample and its augmented versions as a class and learn to group their representations while contrasting with other samples. As a popular loss function in this line, InfoNCE~\cite{oord2018representation} uses multi-class cross-entropy while SimPLE uses binary cross-entropy.

\vspace{2.1mm}

\noindent\textbf{Class proxy design}. Although proxy-based PSL typically updates the class proxies by back-propagation from the loss function, there exist other ways to design class proxies. several methods \cite{hoffer2018fix,pernici2021class,yang2022inducing,Liu2023HUG,li2023no} use fixed classifiers and still obtain satisfactory performance. Liu et al.~\cite{liu2021orthogonal} use stochastic proxies for a large number of categories. Class proxies can also be designed to achieve certain properties~\cite{kasarla2022maximum,mettes2019hyperspherical,pernici2021regular,yang2022inducing} (\eg, uniformity~\cite{liu2018learning,liu2021learning}). Proxy-free methods bypass the difficulty of designing or learning proxies, but they instead introduce the problem of pair construction and mining. Different pair mining strategies in proxy-free PSL may implicitly inject different inductive biases for the learned features, and the mechanism behind is of great importance. How to combine the advantages of proxy-based and proxy-free methods and achieve a good trade-off remains an open challenge.

\vspace{2.1mm}

\noindent\textbf{Deep metric learning}. Traditional metric learning~\cite{xing2002distance,chopra2005learning,hadsell2006dimensionality,weinberger2009distance} learns a proper distance metric that satisfies non-negativity and the triangle inequality. More recently, deep metric learning~\cite{song2016deep,oh2017deep,sohn2016improved} achieves promising performance on image retrieval by using neural networks to learn a feature representation and then put it into a proper distance function. In contrast, PSL drops the requirement to learn a proper distance metric. Recent DML methods~\cite{song2016deep,oh2017deep,sohn2016improved,qian2019softtriple,wang2017deep,wang2019ranked} are mostly triplet-based, while our SimPLE is pair-based.

\vspace{2.1mm}

\noindent\textbf{Concluding remarks}. In this work, we start by rethinking the desideratum of pairwise similarity learning. We then challenge a few common components in current PSL methods, such as angular similarity and margin. We argue that they can be safely removed in pair-based proxy-free frameworks. Following the desideratum, we design a simple yet effective PSL method. Extensive experiments show that SimPLE is able to achieve state-of-the-art performance in a diverse set of open-set recognition tasks.

\vspace{-.75mm}
\section{Limitations and Open Problems} 
\vspace{-1mm}

SimPLE follows a simple yet intuitive design, and yet is by no means, an optimal one. For example, the hard pair mining for positive and negative pairs is still less straightforward and could be further improved. Moreover, similar to existing proxy-free methods, SimPLE can be quite sensitive to the design of pair construction and mining. Despite some limitations, our paper aims to demonstrate that angular similarity is not the only way to achieve state-of-the-art performance.

While SimPLE exhibits empirical superiority in many tasks, a few open problems remain. First, SimPLE introduces three new hyperparameters: $b_{{\theta}}$, $r$, $\alpha$, which means there is one more hyperparameter to tune, compared to the well-known margin-based softmax cross-entropy losses like SphereFace, CosFace, and ArcFace. While hyperparameters in SimPLE have clear physical interpretations, it is still desirable to reduce the number of hyperparameters without sacrificing performance. This requires a deeper understanding of PSL's desideratum. Second, SimPLE achieves less significant performance gain with large training sets. This is mainly due to the naive pair construction strategy, which cannot cover all the representative pairs. We expect that advanced pair sampling methods could be an important future direction. Third, our paper presents a desideratum for general PSL and SimPLE is just one possible route to achieve this desideratum. There may exist many other routes that may potentially perform PSL more effectively. 

Finally, the PSL problem is in fact very general. learning a common embedding space for multi-modal data pairs (\eg, image-text pairs~\cite{radford2021learning}) can also be viewed as a PSL problem. How to apply various PSL methods (including SimPLE) to multi-modal pretraining is a promising direction to explore.

\vspace{-.75mm}
\section*{Acknowledgements}
\vspace{-1mm}

The authors would like to thank colleges from Max Planck Institute for Intelligent Systems at Tübingen for many inspiring discussions. This work was supported by the German Federal Ministry of Education and Research (BMBF): Tübingen AI Center, FKZ: 01IS18039B, and by the Machine Learning Cluster of Excellence, EXC number 2064/1 – Project number 390727645. 

WL was supported by the German Research Foundation (DFG): SFB 1233, Robust Vision: Inference Principles and Neural Mechanisms, TP XX, project number: 276693517. 

AW acknowledges support from a Turing AI Fellowship under grant EP/V025379/1, and the Leverhulme Trust via Leverhulme Centre for the Future of Intelligence.

MJB has received research gift funds from Adobe, Intel, Nvidia, Meta/Facebook, and Amazon.MJB has financial interests in Amazon, Datagen Technologies, and Meshcapade GmbH. While MJB is a consultant for Meshcapade, his research in this project was performed solely at, and funded solely by, the Max Planck Society.

{\small
\bibliographystyle{ieee_fullname}
\bibliography{ref}
}

\newpage
\input{appendix.tex}

\end{document}

%% file: experiments.tex
\section{Experiments and Results}
We evaluate SimPLE with multiple open-set recognition problems, including face recognition, image retrieval, and speaker verification. We adopt the standard training and testing protocols, network configurations, and optimization strategy, so that our results can be transparently and fairly compared to previous methods. The detailed experimental settings are given in the corresponding subsections.

\subsection{Open-set Face Recognition}
\noindent\textbf{Experimental setup}. We generally follow the data processing and augmentation strategy from \cite{kim2022adaface}. Specifically, the face images are cropped based on the 5 face landmarks detected by MTCNN \cite{zhang2016joint} or RetinaFace \cite{deng2020retinaface} using similarity transformation. The cropped image is resized to 112$\times$112, and RGB pixels are normalized to [-1, 1]. In training mode, random cropping, rescaling, and photometric jittering are applied to the face images with a probability of 0.2, while horizontally flipping is applied with the probability of 0.5.

We first evaluate the design of SimPLE by performing ablation studies. SFNet-64 \cite{liu2017sphereface} and MS1MV2 \cite{guo2016ms,deng2019arcface} are adopted as the backbone and training set, respectively. The validation set is constructed by combining LFW \cite{huang2007labeled}, AgeDB-30 \cite{moschoglou2017agedb}, CALFW \cite{zheng2017cross}, and CPLFW \cite{zheng2018cross}, containing 12,000 positive and 12,000 negative pairs. SimPLE models are trained with different $r$ and $\alpha$. The equal error rates (EER) and the true positive rates at different false positive rates (TPR@FAR) on the validation set are reported. 

\vspace{2mm}

\noindent\textbf{Ablation: hyperparameter $r$, $\alpha$, and $b_{\theta}$.} Hyperparameter $r$ is used to control the strength of sample mining. When $r=1$, SimPLE is equivalent to vanilla binary cross-entropy. As $r$ increases, SimPLE focuses more on the hard negative pairs. Table~\ref{tab:ablation_ra} shows that SimPLE yields large performance gains when $r>1$ is used. With $r=3$ and $\alpha=0.001$, SimPLE yields 3.23\% EER, which outperforms the best result with $r=1$ (3.72\%) by a considerable margin.

We also perform a similar ablation study with different $b_{\theta}$, and the results are given in the Appendix. In general, $b_{\theta}=0.3$ or $0.4$ works well for all the experiments. We also observe that higher $\alpha$ is usually paired with higher $r$ for the best performance. With optimal $r$ and $\alpha$ pairs, SimPLE performs equally well. Therefore, we fix $r=3$, $\alpha=0.001$, and $b_{\theta}=0.3$ in the following experiments.

\vspace{2mm}

\noindent\textbf{Ablation: score functions.} To investigate the importance of score functions, we run experiments for SimPLE using cosine similarity or generalized inner product (\ie, Eq.~\ref{eq:gen_inner}) as the score function. Experimental results show that the generalized inner product leads to a significantly lower EER over cosine similarity (with optimal hyperparameters), \ie 3.23\% \emph{vs.} 4.81\%. The results suggest that a proper score function plays a key role in the success of SimPLE.

In our early experimentation, we also attempted to incorporate generalized inner product into the proxy-based framework. However, we did not manage to obtain meaningful results (as shown in the Appendix). This further shows that our SimPLE framework is promising in the sense that it can easily adopt various score functions.

\begin{table}[t]
  \centering
  \scriptsize
  \setlength{\tabcolsep}{3pt}
  \renewcommand{\arraystretch}{1.25}
  \begin{tabular}{cc|cccc}
    $r$ & $\alpha$ & EER ($\downarrow$) & TPR@FAR=1e-4 & TPR@FAR=1e-3 & TPR@FAR=1e-2 \\
    \shline
    1 & 0.0002 & 3.98 & 87.88 & 90.56 & 93.78 \\
    1 & 0.0005 & 3.72 & 89.23 & 92.20 & 94.49 \\
    1 & 0.001 & 3.85 & 88.43 & 90.91 & 94.11 \\
    1 & 0.002 & 4.2 & 85.45 & 89.89 & 93.46 \\ 
    \hline
    2 & 0.0005 & 3.35 & 90.5 & 92.38 & \textbf{94.93} \\
    2 & 0.001 & 3.38 & 88.84 & 92.10 & 94.55\\
    2 & 0.002 & 3.34 & 90.36 & 92.25 & 94.61 \\
    \hline
    3 & 0.0005 & 3.38 & 89.80 & 92.00 & 94.81 \\
    3 & 0.001 & 3.28 & \textbf{91.07} & \textbf{92.45} & 94.80 \\
    3 & 0.002 & \textbf{3.23} & 89.62 & 92.27 & 94.84 \\
  \end{tabular} 
  \vspace{-1.5mm}
  \caption{\small Ablation study of $r$ and $\alpha$ for SimPLE (\%).}\label{tab:ablation_ra}
\vspace{-3mm}
\end{table}

\vspace{1mm}

\noindent\textbf{Comparison with previous methods.} 
For a comprehensive comparison, we conduct experiments under three different settings: (A) SFNet-20 trained with VGGFace2 dataset \cite{cao2018vggface2} (8.6K subjects), (B) SFNet-64 trained with MS1MV2 dataset (85.7K subjects), and (C) IResNet-100 trained with MS1MV2 dataset. The goal is to explore SimPLE under different network capacities and data scales. The evaluations are performed on IARPA Janus Benchmark (IJB) \cite{whitelam2017iarpa, maze2018iarpa}. This is a challenging dataset since it contains mixed-quality samples, \eg low-quality video frames from surveillance cameras and high-quality images. For setting A and B, we train the face models of different methods using their released code, which ensures all methods use the same training recipes except loss functions. For setting C, we directly use the released models or results reported in their published papers, since they represent the current best performance.

\begin{table*}[!t]
  \centering
  \scriptsize
  \setlength{\tabcolsep}{8.75pt}
  \renewcommand{\arraystretch}{1.25}
  \begin{tabular}{l|cccccc|cccccc}
     & \multicolumn{6}{c|}{IJB-B} & \multicolumn{6}{c}{IJB-C} \\
     & \multicolumn{3}{c}{1:1 Verification TAR @ FAR} & \multicolumn{3}{c|}{1:N Identification TPIR @ FPIR} & \multicolumn{3}{c}{1:1 Verification TAR @ FAR} & \multicolumn{3}{c}{1:N Identification TPIR @ FPIR} \\
    Method & 1e-6 & 1e-5 & 1e-4 & top 1 & 1e-2 & 1e-1 & 1e-6 & 1e-5 & 1e-4 & top 1 & 1e-2 & 1e-1 \\
    \shline
    NormFace \cite{wang2017normface} & 32.53 & 68.20 & 82.24 & 91.17 & 58.85 & 78.99 & 65.64 & 76.31 & 86.15 & 92.09 & 70.60 & 81.43 \\
    SphereFace \cite{liu2017sphereface,Liu2021SphereFaceR} & 40.11 & 75.44 & 87.43 & 92.97 & 67.70 & 84.87 & 73.79 & 83.02 & 90.37 & 94.19 & 78.18 & 86.90 \\
    CosFace \cite{wang2018additive,wang2018cosface} & 40.77 & 73.66 & 85.51 & 91.96 & 67.97 & 82.77 & 70.43 & 80.21 & 88.75 & 93.09 & 75.36 & 84.90 \\
    ArcFace \cite{deng2019arcface} & 40.15 & 76.52 & 87.50 & 92.26 & 70.25 & 85.02 & 74.32 & 82.49 & 90.17 & 93.79 & 78.22 & 86.71 \\
    Circle Loss \cite{sun2020circle} & 36.56 & 72.81 & 86.51 & 91.41 & 65.58 & 83.73 & 69.69 & 80.66 & 89.67 & 92.96 & 75.41 & 85.63 \\
    CurricularFace \cite{huang2020curricularface} & 22.16 & 63.35 & 88.23 & 92.66 & 47.59 & 84.93 & 35.54 & 76.49 & 91.10 & 93.73 & 54.13 & 85.77 \\
    SphereFace2 \cite{wen2021sphereface2} & 40.19 & 77.13 & 87.95 & 92.36 & 72.14 & 87.32 & 75.38 & 83.38 & 90.82 & 93.24 & 80.03 & 87.54 \\
    \rowcolor{Gray}
    SimPLE (cosine) & 40.90 & 63.09 & 80.86 & 91.02 & 58.06 & 76.94 & 51.72 & 69.49 & 84.40 & 92.22 & 62.10 & 77.92 \\ 
    \rowcolor{Gray}
    SimPLE & \textbf{47.00} & \textbf{84.51} & \textbf{90.72} & \textbf{93.19} & \textbf{80.44} & \textbf{89.18} & \textbf{82.34} & \textbf{88.62} & \textbf{92.92} & \textbf{94.51} & \textbf{85.66} & \textbf{90.84} \\
  \end{tabular}
  \vspace{-1.5mm}
  \caption{\small Comparison on IJB-B and IJB-C. We use SFNet-20 as the backbone architecture and VGGFace2 as the training set. Results are in \% and higher number indicates better performance.}\label{tab:ijb_sf20}
\end{table*}

\begin{table*}[!t]
  \centering
  \scriptsize
  \setlength{\tabcolsep}{8.75pt}
  \renewcommand{\arraystretch}{1.25}
  \vspace{-.25mm}
  \begin{tabular}{l|cccccc|cccccc}
     & \multicolumn{6}{c|}{IJB-B} & \multicolumn{6}{c}{IJB-C} \\
     & \multicolumn{3}{c}{1:1 Verification TAR @ FAR} & \multicolumn{3}{c|}{1:N Identification TPIR @ FPIR} & \multicolumn{3}{c}{1:1 Verification TAR @ FAR} & \multicolumn{3}{c}{1:N Identification TPIR @ FPIR} \\
    Method & 1e-6 & 1e-5 & 1e-4 & top 1 & 1e-2 & 1e-1 & 1e-6 & 1e-5 & 1e-4 & top 1 & 1e-2 & 1e-1 \\
    \shline
    NormFace \cite{wang2017normface} & 40.56 & 75.30 & 90.22 & 92.49 & 64.62 & 88.19 & 70.17 & 85.88 & 92.69 & 93.70 & 77.97 & 89.81 \\
    SphereFace \cite{liu2017sphereface,Liu2021SphereFaceR} & \textbf{48.83} & 86.66 & 94.36 & 94.84 & 76.35 & 93.20 & 83.57 & 92.79 & 95.82 & 96.07 & 87.74 & 94.47 \\
    CosFace \cite{wang2018additive,wang2018cosface} & 37.82 & 82.99 & 94.20 & 94.69 & 70.61 & 93.03 & 78.01 & 92.29 & 95.87 & 95.91 & 84.59 & 94.53 \\
    ArcFace \cite{deng2019arcface} & 41.02 & 86.16 & \textbf{94.82} & 94.88 & 77.92 & \textbf{93.79} & 84.47 & 93.25 & \textbf{96.25} & 96.12 & 88.80 & \textbf{95.08} \\
    Circle Loss \cite{sun2020circle} & 41.65 & 82.76 & 94.09 & 94.64 & 74.63 & 92.83 & 81.18 & 91.59 & 95.83 & 95.77 & 84.56 & 94.15 \\
    CurricularFace \cite{huang2020curricularface} & 43.76 & 85.55 & 94.61 & 94.82 & 76.01 & 93.37 & 83.35 & 92.95 & 96.11 & 96.04 & 87.88 & 94.76 \\
    SphereFace2 \cite{wen2021sphereface2} & 40.31 & 85.89 & 94.04 & 94.59 & 78.05 & 93.02 & 84.60 & 92.37 & 95.74 & 95.81 & 88.87 & 94.52 \\
    \rowcolor{Gray}
    SimPLE & 46.67 & \textbf{90.34} & 94.49 & \textbf{95.15} & \textbf{83.56} & 93.62 & \textbf{88.49} & \textbf{93.48} & 95.91 & \textbf{96.36} & \textbf{91.88} & 94.76 \\
   \end{tabular}
   \vspace{-1.5mm}
   \caption{\small Comparison on IJB-B and IJB-C. We use SFNet-64 as the backbone architecture and MS1MV2 as the training set.}\label{tab:ijb_sf64}
\end{table*}

\begin{table*}[!t]
  \centering
  \scriptsize
  \setlength{\tabcolsep}{8.75pt}
  \renewcommand{\arraystretch}{1.25}
  \vspace{-.25mm}
  \begin{tabular}{l|cccccc|cccccc}
     & \multicolumn{6}{c|}{IJB-B} & \multicolumn{6}{c}{IJB-C} \\
     & \multicolumn{3}{c}{1:1 Verification TAR @ FAR} & \multicolumn{3}{c|}{1:N Identification TPIR @ FPIR} & \multicolumn{3}{c}{1:1 Verification TAR @ FAR} & \multicolumn{3}{c}{1:N Identification TPIR @ FPIR} \\
    Method & 1e-6 & 1e-5 & 1e-4 & top 1 & 1e-2 & 1e-1 & 1e-6 & 1e-5 & 1e-4 & top 1 & 1e-2 & 1e-1 \\
    \shline
    SphereFace \cite{liu2017sphereface,Liu2021SphereFaceR} & 47.33 & 90.14 & 94.87 & 95.13 & 82.57 & 94.30 & 87.86 & 94.36 & 96.25 & 96.45 & 91.68 & 95.36 \\
    CosFace \cite{wang2018additive,wang2018cosface} & 43.67 & 88.83 & 95.23 & 95.35 & 80.50 & 94.49 & 85.29 & 94.33 & 96.62 & 96.53 & 90.69 & 95.61 \\
    ArcFace ~\cite{deng2019arcface} & 43.43 & 90.40 & 95.02 & 95.14 & 81.36 & 94.26 & 86.00 & 94.49 & 96.39 & 96.47 & 91.91 & 95.51 \\
    CurricularFace$^{\dagger}$ \cite{huang2020curricularface} & - & - & 94.86 & - & - & - & - & - & 96.15 & - & - & - \\
    BroadFace$^{\dagger}$ \cite{kim2020broadface} & 40.92 & 89.97 & 94.97 & - & - & - & 85.96 & 94.59 & 96.38 & - & - & - \\
    SCF-ArcFace$^{\dagger}$ \cite{li2021spherical} & - & 90.68 & 94.74 & - & - & - & - & 94.04 & 96.09 & - & - & - \\
    SphereFace2 \cite{wen2021sphereface2} & 41.53 & 89.92 & 95.02 & 95.24 & 83.46 & 94.36 & 87.63 & 94.49 & 96.42 & 96.41 & 92.08 & 95.47 \\
    MagFace+ \cite{meng2021magface} & 42.32 & 90.36 & 94.51 & 94.81 & 83.65 & 93.87 & 90.24 & 94.08 & 95.97 & 96.02 & 91.95 & 95.06 \\
    AdaFace ~\cite{kim2022adaface} & 46.78 & 90.04 & \textbf{95.67} & \textbf{95.54} & 80.73 & \textbf{95.07} & 89.74 & \textbf{94.87} & \textbf{96.89} & 96.75 & 92.12 & \textbf{96.20} \\
    \rowcolor{Gray}
    SimPLE & \textbf{49.87} & \textbf{91.13} & 94.78 & \textbf{95.54} & \textbf{85.92} & 94.28 & \textbf{90.30} & 94.34 & 96.27 & \textbf{96.81} & \textbf{92.88} & 95.49 \\
  \end{tabular}
  \vspace{-1.25mm}
  \caption{\small Comparison on IJB-B and IJB-C. We use IResNet-100 as the backbone architecture and MS1MV2 as the training set. '-' indicates that neither the model is released nor the result is reported in their paper. $^{\dagger}$ Results are obtained from their papers.}\label{tab:ijb_r100}
\end{table*}

\vspace{1mm}

\noindent\textbf{Setting A: small model and training set.} We first explore SimPLE in a relatively lightweight setting. As can be seen from Table \ref{tab:ijb_sf20}, SimPLE outperforms all competitors by large margins in both verification and identification tasks. In particular, SimPLE respectively outperforms SphereFace2 by 7.38\% and 8.30\% in TAR@FAR=1e-5 and TPIR@FPIR=1e-2 on IJB-B dataset. Similar performance gains can also be observed on the IJB-C dataset, and it shows that SimPLE is effective for low-capacity architectures and small-scale training sets. Using cosine similarity as score function, SimPLE yields inferior results, which is consistent with the performance on the validation set. The results indicate that a lot more small insights (\eg margin) are required before it can achieve competitive performance.

\vspace{2mm}

\noindent\textbf{Setting B and C: larger model and training set.} These experiments are designed to investigate if SimPLE can benefit from larger models and training sets. Again, the comparison is conducted on the IJB datasets and the results are given in Table \ref{tab:ijb_sf64} and Table \ref{tab:ijb_r100}. We observe that SimPLE achieves competitive results on IJB datasets under both settings. Compared to other methods, SimPLE improves more at low accept rates, \eg FPR=1e-6, 1e-5, and FPIR=1e-2. The results validate that SimPLE can benefit from a stronger backbone and more training data. 

\noindent\textbf{Proxy-based vs Proxy-free.} Both SphereFace2 and SimPLE are pair-wise learning frameworks, while SimPLE removes the proxy, angular assumption, and margin term. As shown in Tables \ref{tab:ijb_sf64} and 
\ref{tab:ijb_r100}, the improvement of SimPLE over SphereFace2 suggests that these dominating components might not be necessary in the open-set recognition problem. We hope this observation  will encourage researchers to rethink the use of each component in the PSL framework. 

\begin{table}[!t]
  \centering
  \scriptsize
  \setlength{\tabcolsep}{5.5pt}
  \renewcommand{\arraystretch}{1.25}
  \begin{tabular}{l|ccccc}
    Method & LFW & AgeDB & CALFW & CPLFW & CFP-FP \\
    \shline
    SphereFace \cite{liu2017sphereface,Liu2021SphereFaceR} & 99.78 & 98.02 & 95.56 & 92.11 & 98.08  \\
    CosFace \cite{wang2018additive,wang2018cosface} & 98.81 & 98.11 & 95.76 & 92.28 & 98.12 \\
    ArcFace ~\cite{deng2019arcface} & 98.83 & 98.28 & 95.45 & 92.08 & 98.27 \\
    CurricularFace \cite{huang2020curricularface} & 99.80 & 98.32 & 96.20 & 93.13 & 98.37 \\
    BroadFace \cite{kim2020broadface} & \textbf{99.85} & \textbf{98.38} & 96.20 & 93.17 & 98.63 \\
    SCF-ArcFace \cite{li2021spherical} & 99.82 & 98.30 & 96.12 & 93.16 & 98.40 \\
    SphereFace2 \cite{wen2021sphereface2} & 99.80 & 98.07 & 95.38 & 92.20 & 98.15 \\
    MagFace \cite{meng2021magface} & 99.83 & 98.17 & 96.15 & 92.87 & 98.46 \\
    AdaFace ~\cite{kim2022adaface} & 99.82 & 98.05 & 96.08 & 93.53 & 98.49 \\
    \rowcolor{Gray}
    SimPLE & 99.78 & 98.28 & \textbf{96.25} & \textbf{94.00} & \textbf{98.77} \\
  \end{tabular}
  \vspace{-1.5mm}
  \caption{\small Comparison on multiple high-quality face datasets. Results are in \% and higher number indicates better performance.}\label{tab:val_set}
\end{table}

We further evaluate our SimPLE model trained with setting C on several high-quality datasets, as given in Table \ref{tab:val_set}. SimPLE achieves the highest accuracies on cross-age and cross-pose datasets, \ie 96.25\% on CALFW, 94.00\% on CPLFW, and 98.77\% on CFP-FP, showing the robustness of SimPLE to varying age and pose. The best performance on the LFW and AgeDB datasets (99.85\% and 98.38\%) is obtained by BroadFace, which is a hybrid method that combines proxy-based and proxy-free PSL. Our results suggest 
that the proxy-free PSL paradigm is still worth exploring and should not be ignored for open-set recognition.

\subsection{Image Retrieval and Speaker Verification}

\begin{table}[!t]
  \centering
  \scriptsize
  \setlength{\tabcolsep}{9.5pt}
  \renewcommand{\arraystretch}{1.25}
  \begin{tabular}{l|ccc}
    Method & Precision@1 & R-Precision & MAP@R \\
    \shline
    Contrastive \cite{hadsell2006dimensionality}  & 68.13 & 37.24 & 26.53 \\
    Triplet \cite{weinberger2009distance} & 64.24 & 34.55 & 23.69 \\
    NT-Xent \cite{sohn2016improved,oord2018representation,chen2020simple} & 66.61 & 35.96 & 25.09 \\
    ProxyNCA \cite{movshovitz2017no} & 65.69 & 35.14 & 24.21 \\
    Margin \cite{wu2017sampling} & 63.60 & 33.94 & 23.09 \\
    Margin/class \cite{wu2017sampling} & 64.37 & 34.59 & 23.71 \\
    N. Softmax \cite{wang2017normface, zheng2019hardness} & 65.65 & 35.99 & 25.25 \\
    CosFace \cite{wang2018additive, wang2018cosface} & 67.32 & 37.49 & 26.70 \\
    ArcFace \cite{deng2019arcface} & 67.50 & 37.31 & 26.45 \\
    FastAP \cite{cakir2019deep} & 63.17 & 34.20 & 23.53 \\
    SNR \cite{yuan2019signal} & 66.44 & 36.56 & 25.75 \\
    MS \cite{wang2019multi} & 65.04 & 35.40 & 24.70 \\
    MS+Miner \cite{wang2019multi} & 67.73 & 37.37 & 26.52 \\
    SoftTriple \cite{qian2019softtriple} & 67.27 & 37.34 & 26.51 \\
    \rowcolor{Gray}
    SimPLE & \textbf{68.58} & \textbf{37.62} & \textbf{26.84} \\
  \end{tabular}
  \vspace{-1.5mm}
  \caption{\small Performance of Image Retrieval on CUB-200-2011.}\label{tab:cub2011}
  \vspace{-2mm}
\end{table}

We evaluate SimPLE on two more open-set recognition problems: image retrieval and speaker verification.

\vspace{2mm}

\noindent\textbf{Image Retrieval.} We use the codebase in \cite{musgrave2020metric}, which is a well-known benchmarking toolkit for image retrieval and metric learning. For all the methods, the data processing, training recipes, and testing protocols are nearly the same, except the loss functions. This ensures a fair comparison of different methods. As suggested in \cite{musgrave2020metric}, we use BN-Inception as the backbone \cite{ioffe2015batch} with ImageNet pretraining. The precision at 1 (also known as top-1 / rank-1 accuracy), R-precision, and Mean Average Precision at R (MAP@R) on CUB-200-2011 dataset \cite{wah2011caltech} are reported in Table \ref{tab:cub2011}.

\vspace{2mm}

\noindent\textbf{Speaker Verification.} We adopt the standard train/val/test split given by VoxCeleb2 \cite{chung2018voxceleb2}. The speech recordings are randomly cropped to 3-8 seconds in each mini-batch as data augmentation. The mini-batch size is set to 512. We use ResNet-34 as the backbone architecture. To learn the networks from scratch, the SGD optimizer is used and the learning rate is initialized at 0.1 and divided by 10 after 30K, 50K, and 60K iterations. The training is completed at 70K iterations. We report the EER on VoxCeleb1, VoxCeleb1-easy, VoxCeleb1-hard in Table \ref{tab:voxceleb1}.

\begin{table}[!t]
  \centering
  \scriptsize
  \setlength{\tabcolsep}{8pt}
  \renewcommand{\arraystretch}{1.25}
  \begin{tabular}{l|ccc}
    Method & VoxCeleb1 & VoxCeleb1-E & VoxCeleb1-H \\
    \shline
    Softmax & 2.11 & 2.05 & 3.76 \\
    A-Softmax \cite{liu2017sphereface,Liu2021SphereFaceR} & 2.11 & 2.11 & 3.47 \\
    AM-Softmax \cite{wang2018additive,wang2018cosface} & 2.17 & 2.16 & 3.49 \\
    AAM-Softmax ~\cite{deng2019arcface} & 2.22 & 2.21 & 3.55 \\
    \rowcolor{Gray}
    SimPLE & \textbf{1.85} & \textbf{1.80} & \textbf{3.23} \\
  \end{tabular}
  \vspace{-1.5mm}
  \caption{\small Performance of Speaker Verification on VoxCeleb1.}\label{tab:voxceleb1}
  \vspace{-2.6mm}
\end{table}

Unsurprisingly, SimPLE achieves consistently competitive results on CUB-200-2011 and VoxCeleb1 datasets (Table \ref{tab:cub2011} and \ref{tab:voxceleb1}). The pipeline of different methods is the same, so the gains can only be attributed to the better PSL loss function. This shows that the applications of SimPLE are not limited to any particular object (face) or data modal (image). 
It appears to perform well on a variety 
of open-set recognition problems, \eg generic object or speech data.

%% file: appendix.tex
\onecolumn
\begin{appendix}
\begin{center}


{\LARGE \textbf{Appendix}}
\end{center}

\section{Effect of Hyperparameters}
\label{ablation_btheta}

We explore the effect of $b_{\theta}$ under different hyperparameter settings. The results are given in Table~\ref{tab:ablation_btheta}.

\begin{table}[h]
  \centering
  \scriptsize
  \setlength{\tabcolsep}{3pt}
  \renewcommand{\arraystretch}{1.25}
  \begin{tabular}{ccc|cccc}
    $b_{\theta}$ & $r$ & $\alpha$ & EER ($\downarrow$) & TPR@FAR=1e-4 & TPR@FAR=1e-3 & TPR@FAR=1e-2 \\
    \shline
    0.2 & 1 & 0.0005 & 4.58 & 86.52 & 90.58 & 93.55 \\
    0.2 & 1 & 0.001 & 4.66 & 83.78 & 90.53 & 93.64 \\
    0.2 & 1 & 0.002 & 4.66 & 83.24 & 90.40 & 93.36 \\
    0.2 & 2 & 0.0001 & 3.93 & 88.93 & 91.63 & 94.17 \\
    0.2 & 2 & 0.0002 & 4.29 & 87.55 & 91.03 & 93.57 \\
    0.2 & 2 & 0.0005 & 3.65 & 88.75 & 92.15 & 94.24 \\
    0.2 & 2 & 0.001 & 4.64 & 79.22 & 88.54 & 92.87 \\
    0.2 & 2 & 0.002 & 4.13 & 87.93 & 91.64 & 94.16 \\
    0.2 & 2 & 0.005 & 4.23 & 84.09 & 90.19 & 93.57 \\
    0.2 & 3 & 0.0002 & 6.24 & 73.60 & 82.07 & 89.84 \\
    0.2 & 3 & 0.0005 & 4.07 & 82.18 & 90.20 & 93.28 \\
    \hline
    0.3 & 1 & 0.0002 & 3.98 & 87.88 & 90.56 & 93.78 \\
    0.3 & 1 & 0.0005 & 3.72 & 89.23 & 92.20 & 94.49 \\
    0.3 & 1 & 0.001 & 3.85 & 88.43 & 90.91 & 94.11 \\
    0.3 & 1 & 0.002 & 4.20 & 85.45 & 89.89 & 93.46 \\ 
    0.3 & 2 & 0.0005 & 3.35 & 90.5 & 92.38 & {94.93} \\
    0.3 & 2 & 0.001 & 3.38 & 88.84 & 92.10 & 94.55\\
    0.3 & 2 & 0.002 & 3.34 & 90.36 & 92.25 & 94.61 \\
    0.3 & 3 & 0.0005 & 3.38 & 89.80 & 92.00 & 94.81 \\
    0.3 & 3 & 0.001 & 3.28 & {91.07} & {92.45} & 94.80 \\
    0.3 & 3 & 0.002 & {3.23} & 89.62 & 92.27 & 94.84 \\
    \hline
    0.4 & 1 & 0.0002 & 3.59 & 85.80 & 91.68 & 94.45 \\
    0.4 & 1 & 0.0005 & 3.77 & 85.84 & 91.35 & 94.56 \\
    0.4 & 1 & 0.001 & 3.71 & 87.78 & 91.40 & 94.36 \\
    0.4 & 2 & 0.0005 & 3.58 & 88.50 & 92.33 & 94.94 \\
    0.4 & 2 & 0.001 & 3.34 & 88.67 & 92.85 & 94.94 \\
    0.4 & 2 & 0.002 & 3.35 & 89.25 & 92.12 & 94.50 \\
    0.4 & 3 & 0.0005 & 3.38 & 90.31 & 92.10 & 94.69 \\
  \end{tabular} 
  \vspace{-1.5mm}
  \caption{\small Ablation study of $b_{\theta}$, $r$ and $\alpha$ for SimPLE (\%).}\label{tab:ablation_btheta}
\vspace{-1mm}
\end{table}

There are generally two types of hyperparameters in SimPLE: (1) common hyperparameters which are already extensively studied in existing methods, \eg, learning rate, weight decay, size of the memory buffer; (2) our own hyperparameters: $b_{{\theta}}$, $r$, $\alpha$. \looseness=-1

First, SimPLE only has one more hyperparameter than popular margin-based softmax cross-entropy losses. 
Second, all these hyperparameters are physically interpretable and easy to tune (their feasible range is small). $\thickmuskip=2mu \medmuskip=2mu \alpha\in(0,1)$ is the balancing ratio between positive and negative pairs, which can be set roughly according to the actual ratio of positive and negative pairs in each batch. $\thickmuskip=2mu \medmuskip=2mu b_{\theta}\in(0,1)$ is the angular bias. Setting $b_{\theta}$ around $0.3$ works well. 
$\thickmuskip=2mu \medmuskip=2mu r>0$ controls the easy/hard sample mining. When $r$ gets larger, the loss focuses more on easy positive pairs and hard negative pairs. We empirically show that 
$\thickmuskip=2mu \medmuskip=2mu r=1$ can already achieve good results. More importantly, SimPLE demonstrates consistent performance gain across a wide range of hyperparameter setup (see appendix). The best hyperparameter setup is generally robust to different tasks, datasets, and architectures.\looseness=-1

\newpage
\section{Applying Generalized Inner Product to Proxy-based Methods}
Since generalized inner product achieves significant improvement in SimPLE, we are interested in how it works in proxy-based methods. Here we apply generalized inner product to two representative proxy-based methods: vanilla cross-entropy loss and CosFace. The formulations are given as follows. Note that $\mathcal{L}_{\text{CosFace}^*}$ is equivalent to $\mathcal{L}_{\text{CE}^*}$ when margin $m$ is 0.

\begin{equation}
\small
\mathcal{L}_{\text{CE}^{*}}=\log\big(1+\sum_{i\neq y}\exp(\|\bm{w}_i\|\cdot\|\tilde{\bm{x}}\|\cdot(\cos(\theta_{\tilde{\bm{w}}_i,\tilde{\bm{x}}})-b_{\theta}) - \|\bm{w}_y\|\cdot\|\tilde{\bm{x}}\|\cdot(\cos(\theta_{\tilde{\bm{w}}_y,\tilde{\bm{x}}})-b_{\theta})\big)
\end{equation}
\begin{equation}
\small
\mathcal{L}_{\text{CosFace}^{*}}=\log\big(1+\sum_{i\neq y}\exp(\|\bm{w}_i\|\cdot\|\tilde{\bm{x}}\|\cdot(\cos(\theta_{\tilde{\bm{w}}_i,\tilde{\bm{x}}})-b_{\theta} - m) - \|\bm{w}_y\|\cdot\|\tilde{\bm{x}}\|\cdot(\cos(\theta_1{\tilde{\bm{w}}_y,\tilde{\bm{x}}})-b_{\theta})\big)
\end{equation}

We adopt Setting A in this ablation. $b_{\theta}$ is set from 0 to 0.9 for the vanilla cross-entropy loss. As shown in Table \ref{tab:gip_proxybased}, we do not observe improved performance when applying generalized inner product to the proxy-based method. In contrast, our proxy-free SimPLE can enjoy accuracy gains from the generalized inner product, producing substantially better results.

For CosFace, we also try different combinations of  $b_{\theta}$ and $m$. However, the models do not converge even if we use a very small $m$ (\ie 0.1), resulting in chance-level performance. This is also consistent with the observations in a large body of literature, where cosine similarity has become a \emph{de facto} choice.

\begin{table}[h]
  \centering
  \scriptsize
  \setlength{\tabcolsep}{3pt}
  \renewcommand{\arraystretch}{1.25}
  \begin{tabular}{cc|cccc}
    $b_{\theta}$ & m & EER ($\downarrow$) & TPR@FAR=1e-4 & TPR@FAR=1e-3 & TPR@FAR=1e-2 \\
    \shline
    0 & 0 & 6.12 & 19.12 & 38.23 & 77.82 \\
    0.1 & 0 & 6.08 & 15.48 & 37.95 & 80.70 \\
    0.2 & 0 & 5.99 & 24.11 & 52.16 & 82.88 \\
    0.3 & 0 & 6.10 & 21.02 & 52.77 & 84.27 \\
    0.4 & 0 & 6.40 & 31.58 & 52.03 & 84.30 \\
    0.5 & 0 & 7.00 & 38.78 & 63.12 & 83.08 \\
    0.6 & 0 & 7.47 & 35.78 & 60.34 & 82.07 \\
    0.7 & 0 & 8.05 & 40.47 & 55.13 & 80.58 \\
    0.8 & 0 & 8.91 & 40.33 & 54.98 & 77.36 \\
    0.9 & 0 & 9.83 & 34.63 & 54.72 & 75.23 \\
    0-0.9 & 0.1 & \multicolumn{4}{c}{Not converged} \\
    0-0.9 & 0.2 & \multicolumn{4}{c}{Not converged} \\
    \multicolumn{2}{c|}{CE Loss} & 6.13 & 37.39 & 58.05 & 84.78 \\\hline
    \rowcolor{Gray}
    \multicolumn{2}{c|}{SimPLE (cosine)} & 5.15 & 61.60 & 73.04 & 87.42 \\
  \rowcolor{Gray}
  \multicolumn{2}{c|}{SimPLE} & 4.61 & 68.92 & 78.97 & 90.10 \\
  \end{tabular} 
  \vspace{-1.5mm}
  \caption{\small Results of applying generalized inner product to proxy-based methods.}\label{tab:gip_proxybased}
\vspace{-3mm}
\end{table}

\end{appendix}